\pdfoutput=1
\documentclass[11pt]{article}

\usepackage[margin=0.9in]{geometry}
\usepackage[T1]{fontenc}
\usepackage[utf8]{inputenc}
\usepackage{lmodern}
\usepackage{microtype}
\usepackage{setspace}
\usepackage{amsmath,amssymb,amsthm}
\usepackage{graphicx}
\usepackage{float}
\usepackage{booktabs,tabularx,longtable,array}
\newcommand{\stack}[2][c]{{\renewcommand{\arraystretch}{1.0}\begin{tabular}[t]{@{}#1@{}}#2\end{tabular}}}
\usepackage{caption}
\usepackage{titlesec}
\usepackage{enumitem}
\PassOptionsToPackage{hyphens}{url}
\usepackage[super,numbers,sort&compress]{natbib}
\usepackage[colorlinks=true,linkcolor=black,urlcolor=blue,citecolor=black]{hyperref}

\titleformat*{\section}{\Large\bfseries\sffamily}
\titleformat*{\subsection}{\large\bfseries\sffamily}
\titleformat*{\subsubsection}{\normalsize\bfseries\sffamily}
\titlespacing*{\section}{0pt}{2.2ex plus 1ex minus .2ex}{1.1ex plus .2ex}
\titlespacing*{\subsection}{0pt}{2ex plus 1ex minus .2ex}{0.8ex plus .2ex}
\titlespacing*{\subsubsection}{0pt}{1.8ex plus 1ex minus .2ex}{0.6ex plus .2ex}
\setlist{itemsep=2pt,topsep=4pt}

\makeatletter
\def\@maketitle{%
  \newpage\null\vskip 0.5em
  \begin{center}
    {\LARGE\bfseries\sffamily\@title\par}
    \vskip 1em
    {\large\@author\par}
  \end{center}
  \par\vskip 1.2em}
\makeatother

\AtBeginEnvironment{table}{\small\renewcommand{\arraystretch}{1.15}\setlength{\tabcolsep}{4.5pt}}
\AtBeginEnvironment{longtable}{\scriptsize\renewcommand{\arraystretch}{1.2}\setlength{\tabcolsep}{3pt}}

\title{Spanner: a Weights-Only Lineage Distance for LLMs Across Architectures and Tokenizers}
\author{Matthew Ritch\\\small\href{mailto:matthew.drummond.ritch@gmail.com}{\texttt{matthew.drummond.ritch@gmail.com}}\\\small Code and results: \url{https://github.com/matthew-ritch/spanner}}
\date{}

\begin{document}
\maketitle

\begin{abstract}
Open LLMs are heavily remixed, so their lineages are difficult to trace. Existing weight-based model lineage measurements are restricted to models of the same size or tokenizers or do not properly handle model symmetries. Output-based lineage measurements cannot distinguish weight descent from distillation. We introduce Spanner ($D$), a weights-only pseudometric that projects attention and MLP invariants onto shared tokens and compares layer spans with Yanai's GCD. Spanner compares LLMs across architectures, tokenizers, and dimensions while ignoring symmetries such as residual stream rotation that do not change the models' outputs. We show that $D$ is invariant to model symmetries, that $D^2$ is an effective measurement of cumulative learning rate on Pythia ($R^2 = 0.98$), and that $D$ is competitive with or outperforms published baselines (0.991 AUROC on AWM, 0.889 on LLM Biometrics). We also show that $D$ separates weight descent from behavioral convergence. We use $D$ to audit a set of open models on the Hugging Face Hub and find several models with potentially misattributed parents. Excluding download time, $D$ can calculate a full distance matrix between 1,000 8B models in about 17 hours for \$30 on a Hugging Face l40sx1, so our method enables large-scale open model lineage tracing and auditing.
\end{abstract}

\section{Introduction}
\label{sec:intro}

Most open LLMs are built from other LLMs\citep{huggingface2026models}, but their lineages can be opaque. Developers, researchers, and hobbyists create new models with remixing techniques such as fine-tuning, merging, splitting, splicing, quantizing, upcycling, and applying task vectors\citep{ilharco2023editing,yadav2023tiesmerging,yu2024language,huang2024chat,goddard2024arcees}. LLMs can also be trained via distillation from other LLMs to mimic them without access to their underlying weights\citep{hinton2015distilling}. Models with shared lineage can carry shared capabilities and fine-tuning can change or remove alignment\citep{qi2024finetuning}. However, new LLMs are produced through a combination of methods and their documentation is frequently incomplete, so their lineages can be unclear.

Researchers have built a host of methods to evaluate lineage (Table~\ref{tab:related}). Lineage measurements require various assumptions of model access\citep{shao2025sok}, including access to weights\citep{zeng2026awm}, log-probabilities\citep{ogawa2026reconstructing}, or only generated tokens\citep{yax2025phylolm}, and serve various purposes, including hypothesis testing for independence and creating trees of model lineages. Weight-based LLM distance measurements compare invariants projected onto tokens\citep{zeng2024huref,dar2023analyzing}, matched attention weights\citep{zeng2026awm}, distributions of singular values\citep{wang2026ghost}, weight matrix traces and subspaces\citep{chen2026who}, or use statistics such as weight kurtosis to estimate lineage direction\citep{horwitz2025unsupervised,li2026witness}. Activation or residual-based methods\citep{zhu2025independence,zhang2025reef} compare models' responses to probe prompting, which requires forward passes. Behavioral or outputs-based methods\citep{ogawa2026reconstructing,nikolic2025model,yax2025phylolm,wu2026llm} work across architectures and tokenizers, but they conflate mimicry and shared behavior with weight descent. Activation and behavioral methods require carefully tuned probe input sets.

\begin{table}[H]
\centering
\caption{Related methods. $\checkmark$ yes, $\times$ no, $\sim$ partial invariance or function across tokenizers or model dimensions, -- does not apply.}
\label{tab:related}
\begin{tabular}{@{}lllcccc@{}}
\toprule
Method & Inputs & Output & \stack{Different\\Widths} & \stack{Different\\Depths} & \stack{Different\\Tokenizers} & \stack{Ignore\\Symmetries} \\
\midrule
HuRef\citep{zeng2024huref} & weights & distance & $\checkmark$ & $\times$ & $\sim$ & $\sim$ \\
AWM\citep{zeng2026awm} & weights & distance & $\checkmark$ & $\checkmark$ & $\times$ & $\sim$ \\
GhostSpec\citep{wang2026ghost} & weights & distance & $\sim$ & $\checkmark$ & $\checkmark$ & $\sim$ \\
Biometrics (Trace)\citep{chen2026who} & weights & distance & $\checkmark$ & $\checkmark$ & $\checkmark$ & $\sim$ \\
Biometrics (subspace)\citep{chen2026who} & weights & distance & $\times$ & $\times$ & -- & $\times$ \\
MoTHer\citep{horwitz2025unsupervised} & weights & directed tree & $\times$ & $\times$ & -- & $\times$ \\
Witness Overlap\citep{li2026witness} & weights & direction & $\times$ & $\times$ & -- & $\times$ \\
Zhu et al.\citep{zhu2025independence} & activations & hypothesis test & $\checkmark$ & $\checkmark$ & $\times$ & $\checkmark$ \\
REEF\citep{zhang2025reef} & activations & distance & $\checkmark$ & $\sim$ & $\sim$ & $\checkmark$ \\
TRACE\citep{ogawa2026reconstructing} & log-probs & directed tree & $\checkmark$ & $\checkmark$ & $\sim$ & $\checkmark$ \\
Provenance testing\citep{nikolic2025model} & outputs & hypothesis test & $\checkmark$ & $\checkmark$ & $\checkmark$ & $\checkmark$ \\
PhyloLM\citep{yax2025phylolm} & outputs & tree & $\checkmark$ & $\checkmark$ & $\checkmark$ & $\checkmark$ \\
\textbf{Spanner ($D$)} & weights & distance, tree & $\checkmark$ & $\checkmark$ & $\checkmark$ & $\checkmark$ \\
\bottomrule
\end{tabular}
\end{table}

We propose Spanner, a new method for calculating a distance $D$ between models that fulfills the following design constraints:
\begin{enumerate}
  \item Operate directly on model weights so it is independent of sampling noise, bias from probe prompts, and pairs with outputs-only model lineage methods such as PhyloLM\citep{yax2025phylolm} to separate weight descent from mimicry.
  \item Compare models across widths, depths, and tokenizers so we can capture the full breadth of weight descent possible with model remixing.
  \item Ignore model symmetries that do not change outputs, such as residual basis rotations\citep{elhage2021mathematical,ashkboos2024slicegpt}, rescaling symmetry, and attention head and neuron permutations\citep{ainsworth2023git,theus2025generalized}.
  \item Qualify as a pseudometric between any two LLMs so lineage trees are well defined.
  \item Ensure that duplicating any layer in a model doesn't change its distance to other models.
  \item Enable fast pairwise distance calculation.
\end{enumerate}

Spanner projects model invariants down to the models' shared token space\citep{elhage2021mathematical,zeng2024huref} to compare across architectures. It compares models by their layers' spans with Yanai's GCD\citep{yanai1974unification,ramsay1984matrix,kornblith2019similarity}, ensuring that $D$ is a pseudometric\citep{williams2021generalized}. It uses TensorSketch to minimize the calculation cost of each pairwise comparison\citep{pham2013fast,charikar2002finding}. Spanner does not determine descent direction and cannot detect distillation alone.

We begin with two basic experiments. First, we verify that $D$ is invariant to model symmetries and can detect remixes including splits, merges, and quantization (Section~\ref{sec:exp-invariants}). Then, we use Pythia\citep{biderman2023pythia} to verify ordering over training checkpoints and to assess $D$ as a ``molecular clock'' of training (Section~\ref{sec:exp-clock}). We then compare $D$ to published methods using their benchmarks (Section~\ref{sec:exp-benchmarks}):
\begin{enumerate}
  \item $D$ scores 0.991 AUROC against AWM\citep{zeng2026awm}'s perfect 1.000 AUROC on AWM's own evaluation set. Both outperform REEF\citep{zhang2025reef}, HuRef\citep{zeng2024huref}, and PCS\citep{zeng2024huref}, which are also reported in the AWM paper.
  \item $D$ scores 0.889 AUROC against LLM Biometrics\citep{chen2026who}'s 0.770 AUROC on its most difficult evaluation set. That AUROC is from our rerun, but \citet{chen2026who} report 0.850 AUROC in their paper.
  \item PhyloLM\citep{yax2025phylolm}'s distance is smaller between unrelated models from the same lab and larger between models from different labs because they behave alike, whereas $D$ measures both cases as independent.
\end{enumerate}

We then compare $D$ and PhyloLM's recovery of known model lineages (Section~\ref{sec:exp-lineage}). Finally, we apply $D$ ``in the wild'' to explore Qwen derivatives and a broad inter-lab set of models on Hugging Face (Section~\ref{sec:exp-wild}). We found several potentially misattributed model parents while verifying model lineage there. We show that $D$ can map lineage in 1,000 8B parameter models for \$30 of compute (Section~\ref{sec:exp-cost}). Our code and results are available at \url{https://github.com/matthew-ritch/spanner}.

\section{Methodology}
\label{sec:method}

In this section, we describe the procedure for calculating $D$, show how it meets our design goals, and describe the resulting tradeoffs and limitations. To compute $D$, we describe two models' layers as matrices mapping token space back into token space and then compare their spans. Models that share weight ancestry, such as a fine-tuned model and its base, share spans, while independent models are nearly orthogonal. We will use row vectors for the residual stream.

\subsection{Procedure}
\label{sec:procedure}

\subsubsection{1. Fold linear maps into adjacent matrices}
\label{sec:step-fold}

We would like to compare LLMs modulo rescaling symmetry\citep{dinh2017sharp,ashkboos2024slicegpt}. Transformers apply linear maps such as norm scales, LayerNorm's centering, and constant multiples before and after attention and MLP sublayers. Multiplying each channel of a norm by a positive constant and the matching row of the next matrix by its inverse does not change their product or the model overall.

To remove this ambiguity, we fold each linear map into its adjacent matrix and ignore normalization and biases. For example, if a layer reads from the residual stream through RMSNorm with scale $\gamma$, we fold that into the adjacent matrices and replace $W$ with $\operatorname{diag}(\gamma) W$. In post-norm models, the residual stream norms are folded into each sublayer's read and write bases and for LayerNorm we fold the centering map $C = I - \frac{1}{d} \mathbf{1}\mathbf{1}^\top$ into each read and write matrix.

\subsubsection{2. Token coordinates}
\label{sec:step-tokens}

We would like to compare LLMs across different tokenizers. First, we find the set of tokens $S$ that is present in every model's vocabulary. See Appendix~\ref{app:tokens} for token set sizes by experiment.

Then, take $E \in \mathbb{R}^{|S| \times d}$, the model's embedding rows on $S$, and $U \in \mathbb{R}^{|S| \times d}$, its unembedding rows on $S$, and apply step 1's folds to both $E$ and $U$. This means we fold $\gamma_{\text{final}}$ into $U$, so $U \neq E$ even with tied models. We center both $E$ and $U$ by subtracting their average embedding.

\subsubsection{3. Calculate invariant terms and project onto tokens}
\label{sec:step-invariants}

We would like $D$ to be invariant to attention head and hidden unit permutations and rotations of the residual stream, so we compute a set of $\mathbb{R}^{d \times d}$ invariants for each attention and MLP sublayer\citep{elhage2021mathematical}, then project them onto token coordinates. For attention, there are \citet{elhage2021mathematical}'s QK and OV circuits, which respectively represent a head's selective attention and outputs. We compare models with the attention and MLP sublayers separately and sum across heads and neurons for head-permutation invariance. For attention sublayers, we calculate the following invariants. We index the query heads with $h$. We normalize each $W_{QK}$ to compare token relations without scale.
\[
W_Q^h, W_K^h \in \mathbb{R}^{d \times d_k}, \quad W_V^h \in \mathbb{R}^{d \times d_v}, \quad W_O^h \in \mathbb{R}^{d_v \times d}
\]
\[
Q = \sum_{h} \frac{W_{QK}^h}{\lVert W_{QK}^h \rVert_F}\text{, where } W_{QK}^h = W_Q^h (W_K^h)^\top \quad \text{(QK)}
\]
\[
O = \sum_{h} W_V^h W_O^h \quad \text{(OV)}
\]

For MLP sublayers, we calculate the following invariants. We call neuron $j$'s up vector $u_j$, its output vector $o_j$, and if it has one, its gate vector $g_j$. $G_{\text{in}}$ and $G_{\text{out}}$ encode the MLP sublayer's read and write directions. $M$ encodes each neuron's key-value pair behavior\citep{geva2021transformer}. We will use $\hat{v}=\frac{v}{\lVert v \rVert_2}$. We normalize the vectors in $G_{\text{in}}$ and $G_{\text{out}}$ but keep raw vectors in $M$ so all three are invariant to per-neuron rescaling $u_j \to c_j u_j, o_j \to c_j^{-1} o_j$. For MoE layers, we include all experts' neurons in sums over $j$ and ignore the router.
\[
u_j, g_j, o_j \in \mathbb{R}^d \quad \text{column vectors}
\]
\[
G_{\text{in}} = \sum_{j} ( \hat{u}_j \hat{u}_j^\top + \hat{g}_j \hat{g}_j^\top )
\]
\[
G_{\text{out}} = \sum_{j} \hat{o}_j \hat{o}_j^\top
\]
\[
M = \sum_{j} u_j o_j^\top
\]

We center each invariant $F \leftarrow F - \frac{\operatorname{tr}(F)}{d}I$ and use $U$ and $E$ to project each invariant onto an $|S| \times |S|$ matrix over $S$. We use $T$ as a generic variable for each projected matrix.
\begin{itemize}
  \item $O$ and $M$: $T = E F U^\top$
  \item $Q$ and $G_{\text{in}}$: $T = E F E^\top$
  \item $G_{\text{out}}$: $T = U F U^\top$
\end{itemize}

\subsubsection{4. Sketch and assemble the projected invariants}
\label{sec:step-sketch}

Each $T \in \mathbb{R}^{|S| \times |S|}$. In realistic model comparison, $|S|$ can be large. We represent it with a TensorSketch\citep{pham2013fast,charikar2002finding} so we can compute its inner products without forming it as a raw matrix. Sketching changes $D$ by less than $10^{-3}$ for the 1B--2B models tested. We use $k = 16{,}384$ in our reference implementation and for the results in this paper unless otherwise stated. See Appendix~\ref{app:sketch} for details.

We $\ell_2$ normalize each sketch and then concatenate them to form a fingerprint vector $f_l$ for each sublayer.

\subsubsection{5. Yanai's generalized coefficient of determination}
\label{sec:step-gcd}

We want to calculate a distance between models that compares their weights modulo ordering, recombination, and other remixing techniques, so we compare the spans of the two models with Yanai's generalized coefficient of determination\citep{yanai1974unification,ramsay1984matrix,kornblith2019similarity}, which yields $D$, the angle between those two subspaces.
\begin{itemize}
  \item For each model and each sublayer type $t \in \{\text{attention}, \text{MLP}\}$, assemble each $f_l$ into the columns of a matrix $\Phi_t \in \mathbb{R}^{n_t k \times L_t}$ where $L_t$ is the number of sublayers of type $t$ and $n_t$ is 2 for attention layers and 3 for MLP layers.
  \item Define $K_t = \Phi_t^\top \Phi_t$ and compute $K_t^+$ from the eigendecomposition of $K_t$. Drop eigenvalues $\lambda < 10^{-10}\lambda_{\max}$ for stability.
  \item Compute the projector onto the span of $\Phi_t$'s sublayers, $P_t = \Phi_t K_t^+ \Phi_t^\top$.
  \item Compute $\langle P_A, P_B \rangle_t = \left\| (K_A^+)^{1/2} \, (\Phi_A^\top \Phi_B) \, (K_B^+)^{1/2} \right\|_F^2$
  \item Compute $\|P\|_t^2 = r_t$, the rank of $K_t$.
  \item Pool across the layer types and compute $\cos\theta = \dfrac{\langle P_A, P_B \rangle_{\text{attn}} + \langle P_A, P_B \rangle_{\text{MLP}}}{\sqrt{(r_{A,\text{attn}} + r_{A,\text{MLP}})(r_{B,\text{attn}} + r_{B,\text{MLP}})}}$
  \item Finally, compute $\mathbf{D(A, B) = \arccos(\cos\theta)}$
\end{itemize}

This $D$ is the arccos of the normalized Frobenius inner product of the direct sum of the attention and MLP projectors. Thus, it is an angle between unit vectors and is thus a metric\citep{williams2021generalized} between the equivalence classes of LLMs modulo our chosen invariants and a pseudometric between any two LLMs. $D$ is an angle between high-dimensional unit vectors, so independent changes are nearly orthogonal and we can approximate a series of small perturbations with the sum of their squares. For that reason, we use $D^2$ as the additive quantity for tree creation algorithms.

\subsection{Proof of Invariance and Pseudometricity}

\newtheorem{proposition}{Proposition}

\begin{proposition}[Invariance and pseudometricity]
For a fixed token set $S$ and fixed CountSketch hashes, $D$ is:
(i) zero between any two models that differ only by rescaling symmetries, scaled orthogonal rotations of the residual stream ($h \mapsto a h R$ with $a > 0$ and $R^\top R = I$), attention head or MLP neuron ordering, attention head bases ($W_V^h X, X^{-1} W_O^h$, and $W_Q^h Y, W_K^h (Y^{-1})^\top$ in models without QK-norm), or MLP neuron scaling ($u_j \mapsto c_j u_j, o_j \mapsto c_j^{-1} o_j$, $c_j > 0$), and invariant to duplicating or reordering sublayers,
(ii) a pseudometric on models.
\end{proposition}

\begin{proof}[Proof sketch]
(i) Step~1 folds norm scales into their adjacent matrices, removing scale differences across norms. Each invariant $F$ sums over heads or neurons, ignoring permutations, and $X$, $Y$, and $c_j$ cancel inside each invariant. Rotating and scaling the residual stream right-multiplies $E$ and $U$ by $R$ and replaces $F$ with $R^\top F R$ up to a positive scalar. $R$ is orthogonal ($R R^\top = I$), so the rotations cancel in $E F U^\top$, $E F E^\top$, and $U F U^\top$, and any positive scalar cancels when we $\ell_2$-normalize each sketch. Duplicating or reordering sublayers does not change the span of $\Phi_t$, so its projector $P_t$ is also unchanged.
(ii) With fixed and shared token set $S$ and sketch hashes, define $\psi(A) = \frac{P_{A,\mathrm{attn}} \oplus P_{A,\mathrm{MLP}}}{\sqrt{r_{A,\mathrm{attn}} + r_{A,\mathrm{MLP}}}}$, which depends only on model $A$. Because each $P_{A,t}$ is an orthogonal projector of rank $r_{A,t}$, $\|P_{A,t}\|_F^2 = r_{A,t}$ and therefore $\|\psi(A)\|_F = 1$. Take a second model $B$. The Frobenius inner product of direct sums adds, so $\langle \psi(A), \psi(B) \rangle_F = \frac{\langle P_{A,\mathrm{attn}}, P_{B,\mathrm{attn}} \rangle + \langle P_{A,\mathrm{MLP}}, P_{B,\mathrm{MLP}} \rangle}{\sqrt{(r_{A,\mathrm{attn}} + r_{A,\mathrm{MLP}})(r_{B,\mathrm{attn}} + r_{B,\mathrm{MLP}})}} = \cos\theta$, so $D(A,B) = \arccos(\langle \psi(A), \psi(B) \rangle_F)$ is the angle between two unit vectors, which is a metric on unit vectors\citep{williams2021generalized}, and hence a pseudometric on models.
\end{proof}

\section{Experiments}
\label{sec:experiments}

\subsection{Invariants}
\label{sec:exp-invariants}

We designed $D$ to ignore a set of symmetries, so $D$ should be 0 between models that differ only by those symmetries. We also designed it to detect a variety of model remixing techniques. In this experiment, we validate these properties (Table~\ref{tab:invariants}).
\begin{itemize}
  \item We use Qwen3-1.7B (Base and Instruct)\citep{yang2025qwen3} and Llama-3.2-1B(-Instruct)\citep{grattafiori2024llama} as our base models.
  \item We assess $D$ on symmetries including exact copies, random rotation and scaling of the residual stream, head and neuron permutations, and rescaling symmetries.
  \item We also observe how additive weight noise, quantization, and fine-tuning affect $D$.
  \item We verify that $D$ can detect remixing techniques including duplicating and dropping layers, splicing layers from different models, swapping embeddings, merges, and assembling MoEs from fine-tuning.
\end{itemize}

\begin{table}[htbp]
\centering
\caption{Invariants and remixes.}
\label{tab:invariants}
\begin{tabularx}{\textwidth}{@{} >{\raggedright\arraybackslash}X l >{\raggedright\arraybackslash}p{4.8cm} @{}}
\toprule
Edit & Predicted $D$ & Observed $D$ \\
\midrule
8 exact symmetries (Qwen3-1.7B) & 0 & $3.8\times 10^{-7}$ -- $4.3\times 10^{-6}$ \\
9 exact symmetries (Llama-3.2-1B) & 0 & $3.9\times 10^{-7}$ -- $4.6\times 10^{-6}$ \\
Layers 8--15 duplicated & 0 & $4.2\times 10^{-7}$ \\
Layers 20--23 dropped (24 of 28) & $\arccos(\sqrt{24/28}) = 0.38760$ & 0.38760 \\
Gaussian noise from $10^{-4} \to 10^{-1}$ & $D$ increases with $\sqrt{\mathrm{KL}}$ & $\frac{D^2}{\mathrm{KL}} = 0.080\text{--}0.082$ \\
8-Bit Quantization & -- & 0.020 \\
4-Bit Quantization & -- & 0.258 \\
3-Bit Quantization & -- & 0.528 \\
Fine-tuned 10 steps & increases with duration & 0.0046 \\
Fine-tuned 100 steps & increases with duration & 0.0176 \\
Qwen3-1.7B Instruct embeddings on Base & -- & 0.199 (48.1\% of $D^2(\text{Base}, \text{Instruct})$) \\
Llama-3.2-1B Instruct embeddings on Base & -- & 0.337 (48.0\% of $D^2(\text{Base}, \text{Instruct})$) \\
Qwen3-1.7B splice (Instruct layers 0--13 + embeddings, Base layers 14--27) & -- & 0.156 to Instruct, 0.242 to Base \\
Linear 50/50 and SLERP\citep{shoemake1985animating}, Qwen3-1.7B Instruct and Base & -- & 0.143--0.146 to each parent \\
Task arithmetic\citep{ilharco2023editing}, $\lambda = 1.5$ (Qwen3-1.7B Instruct's task vector on Base) & -- & 0.430 to Base, 0.148 to Instruct \\
TIES\citep{yadav2023tiesmerging} (Qwen3-1.7B and three fine-tunes) & -- & 0.183--0.225 to each parent \\
DARE\citep{yu2024language} $p = 0.9$ & -- & 0.77--0.81 to each parent \\
MoE ($4 \times$ Qwen3-0.6B) & small but nonzero & 0.006--0.047 \\
\bottomrule
\end{tabularx}
\end{table}

KL is the KL divergence $\mathrm{KL}(p_{\text{original}} \parallel p_{\text{edited}})$ of next-token distributions in nats averaged over every position of 200 512-token windows of the WikiText-103 test set. We added Gaussian noise relative to each layer matrix's RMS, evaluated on Qwen3-1.7B.

Our design for $D$ meets our stated design goals. As designed, $D$ ignores our chosen symmetries down to rounding error, $D^2 \propto \mathrm{KL}$ over three orders of magnitude of Gaussian noise, and various remixing techniques change $D$ as we expected.

\subsection{Training clock}
\label{sec:exp-clock}

\begin{figure}[htbp]
\centering
\includegraphics[width=\textwidth]{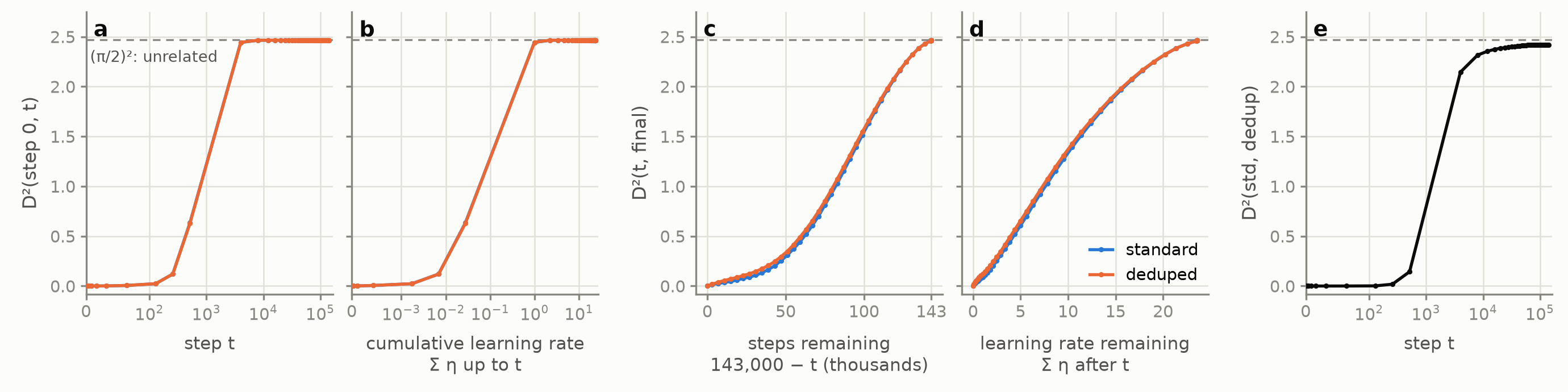}
\caption{$D^2$ vs.\ training step, cumulative learning rate, and cross-run divergence.}
\label{fig:clock}
\end{figure}

How does $D$ change along training trajectories? Pythia\citep{biderman2023pythia} is a well-studied suite of LLMs with publicly available training checkpoints. We evaluate $D$ on Pythia-410M's training trajectories from both their standard and deduplicated training runs, which have nearly identical corpora (Figure~\ref{fig:clock}).

$D$'s behavior on the Pythia suite aligns with our design. $D^2(t, \text{final})$ is monotone and near-linear with the cumulative learning rate remaining at step $t$ with a 0.98 $R^2$ (Figure~\ref{fig:clock}d), so it is an effective measure of cumulative learning rate. This behavior is consistent with previous reports of model changes scaling with cumulative learning rate\citep{tissue2025scaling,schurholt2020investigation}. Additionally, $D(\text{standard}, \text{deduped})$ grows from 0 at step 0 to 0.38 at step 512, 1.46 at step 4,000, and 1.55 at step 143,000 (Figure~\ref{fig:clock}e), indicating that $D$ evaluates shared lineage and not shared corpora.

\subsection{Spanner and published benchmarks}
\label{sec:exp-benchmarks}

How does $D$ compare to published methods with similar purposes on their own benchmarks?

\subsubsection{Comparison with AWM}
\label{sec:exp-awm}

AWM\citep{zeng2026awm} is a weights-only lineage measurement that works across model widths and depths but is restricted to comparisons within a single tokenizer. \citet{zeng2026awm} evaluated AWM alongside various other lineage measurement methods with a set of 60 pairs of related models and 90 pairs of unrelated models. They included 10 related pairs for each of the following categories: fine-tunes, RL, continued pretraining, pruning, MoE upcycling and multimodal models.

We evaluated $D$ and Trace\citep{chen2026who}, but the rest are from \citet{zeng2026awm}'s reported results (Table~\ref{tab:awm}).

\begin{table}[htbp]
\centering
\caption{Comparison with AWM.}
\label{tab:awm}
\begin{tabular}{@{}llcc@{}}
\toprule
Method & Inputs & AUROC & TPR at 1\% FPR \\
\midrule
AWM\citep{zeng2026awm} & weights & 1.000 & 1.000 \\
Spanner ($D$) & weights & 0.991 & 0.967 \\
HuRef\citep{zeng2024huref} & weights & 0.957 & 0.867 \\
Trace\citep{chen2026who} & weights & 0.959 & 0.667 \\
PCS\citep{zeng2024huref} & weights & 0.86 & 0.43 \\
REEF\citep{zhang2025reef} & activations & 0.778 & 0.30 \\
\bottomrule
\end{tabular}
\end{table}

Two of the models in AWM's benchmark have small token sets, so we scored pairs involving those models with per-pair token intersections. \nolinkurl{mosaicml/mpt-7b} was removed from Hugging Face, so we used the mirror \nolinkurl{anas-awadalla/mpt-7b}.

AWM's benchmark was assembled by its authors to evaluate the method. We use these pairs exactly as published, so the comparison may favor AWM. Despite this, our method is competitive with these published results.
\begin{itemize}
  \item $D$ is within 0.01 AUROC of AWM and outperforms the other methods
  \item $D$ misses two related pairs:
  \begin{itemize}
    \item $D(\text{Llama-3.2-1B}, \text{Llama-3.1-8B}) = 1.565$. This relationship is pruning followed by long distillation.
    \item $D(\text{Qwen2.5-7B},\allowbreak\ \text{Nemotron-Research-Reasoning-Qwen-1.5B}) = 1.567$. The benchmark's readme says Nemotron is an RL fine-tune of Qwen2.5-7B, but its model card says DeepSeek-R1-Distill-Qwen-1.5B is its base model. Further, AWM's code pairs Qwen2.5-7B with Open-Reasoner-Zero-7B instead of Nemotron. We score the pair as stated, but we believe the label is incorrect.
  \end{itemize}
\end{itemize}

\subsubsection{Comparison with LLM Biometrics}
\label{sec:exp-biometrics}

In LLM Biometrics, \citet{chen2026who} propose Trace, a weights-only method that is similar to $D$ in that it works across model shapes and tokenizers. They also propose Subspace Alignment, which only works between models of the same shape. In this experiment, we evaluate these two methods against $D$ on their own benchmark as well as on AWM's benchmark (Table~\ref{tab:biometrics}). Their benchmark is organized into three sets of pairs: S1 (independently trained models), S2 (models of the same series, mostly pretrained separately) and S3 (fine-tunes of a shared base). Their benchmark assesses whether a method gives a related pair, such as two models from S2, a lower distance than two unrelated models, such as a pair from S1.

\begin{table}[htbp]
\centering
\caption{Comparison with LLM Biometrics.}
\label{tab:biometrics}
\begin{tabular}{@{}llccc@{}}
\toprule
Benchmark & Metric & Spanner ($D$) & Trace & Subspace \\
\midrule
S1 vs S2 & AUROC & \textbf{0.889} & 0.770 & -- \\
S2 vs S3 & AUROC & 1.000 & 1.000 & -- \\
S1 vs S3 & AUROC & 1.000 & 1.000 & -- \\
AWM, all 150 pairs & AUROC & \textbf{0.991} & 0.959 & -- \\
AWM, 118 same-shape pairs & AUROC & 1.000 & 0.976 & 1.000 \\
\bottomrule
\end{tabular}
\end{table}

Note that we were unable to reproduce Trace's published 0.850 AUROC on S1 vs S2. Subspace can only be calculated between models of the same shape.

$D$ matches or outperforms both methods proposed in \citet{chen2026who} on both their benchmark and AWM's benchmark.

\subsubsection{Comparison with PhyloLM}
\label{sec:exp-phylolm}

PhyloLM\citep{yax2025phylolm} evaluates model lineages and builds trees using model outputs. It samples each model's responses to a set of probe prompts and measures the distance between their outputs. We included this comparison to compare performance between our weights-based $D$ and the outputs-based PhyloLM, but we are also interested in the differences between detections possible with these two approaches. A complete analysis of model descent requires both evaluating weight descent and distillation or mimicry. In this section, we use the open-weight models in PhyloLM's published evaluation set and the completions published with the paper. See the model zoo (Appendix~\ref{app:zoo}) for details. We assembled a documented lineage for these models from their creators' model cards and publications. We grade the methods on their ability to recover these lineages.

We compare PhyloLM and $D$'s ability to organize models into lineages with three-taxon lineage tests\citep{nelson1991threetaxon} (Table~\ref{tab:phylolm-triples}):
\begin{itemize}
  \item Betweenness, where for each method's distance $d$, the method passes if $d(a,b) > d(a,m)$ and $d(a,b) > d(b,m)$:
  \begin{itemize}
    \item Parent of two: $a$ and $b$ descend from $m$ on different branches, for example a base model and two fine-tunes.
    \item Chain: $a \to m \to b$ is a direct line of descent. For example, a base model, its instruct model, and a fine-tune of the instruct model.
    \item Ancestor and outgroup: $a$ descends from $m$, but $b$ is neither an ancestor nor a descendant of $m$.
  \end{itemize}
  \item Siblings closer than cousins: none of the three models is an ancestor of another model, but we know from documented lineage that one pair, $a$ and $b$, is more closely related to each other than they are to the third model, $c$. The method passes if $d(a,c) > d(a,b)$ and $d(b,c) > d(a,b)$.
\end{itemize}

We do not run betweenness tests across lineages because they ask if $m$ is on the lineage path between $a$ and $b$, but if $a$ and $b$ are from different lineages then there is no path between them, so the test has no correct answer. Comparisons of triples within a single lineage are not independent, so we report 95\% confidence intervals from a cluster bootstrap over the 54 documented lineages.

\begin{table}[htbp]
\centering
\caption{Three-taxon lineage tests on PhyloLM's evaluation set.}
\label{tab:phylolm-triples}
\setlength{\tabcolsep}{3.5pt}
\begin{tabular}{@{}lccc@{}}
\toprule
Test & $n$ & Spanner [95\% CI] & PhyloLM [95\% CI] \\
\midrule
Betweenness: ancestor and outgroup & 298 triples & 74.8\% [58.6, 95.5] & 55.0\% [47.8, 82.6] \\
Betweenness: parent of two & 450 triples & 79.1\% [65.6, 84.8] & 66.7\% [26.6, 74.9] \\
Betweenness: chain & 72 triples & 63.9\% [48.4, 100] & 51.4\% [38.5, 83.9] \\
Siblings closer than cousins within one lineage & 332 triples & 91.0\% [89.7, 100] & 88.9\% [87.6, 97.6] \\
Siblings closer than cousins across lineages & 61,526 triples & 99.2\% [95.1, 100] & 97.0\% [87.1, 99.6] \\
\bottomrule
\end{tabular}
\end{table}

We also compared PhyloLM to $D$ on a set of models chosen to see where their results deviate rather than to compare performance (Table~\ref{tab:phylolm-rel}). This set contained fine-tunes, distillations, remixes, and other pairs. See the model zoo (Appendix~\ref{app:zoo}) for details. Note the agreement on fine-tunes and merges and the disagreement on distillation and unrelated models from the same lab. We report medians and 95\% CIs of the median because PhyloLM's results are skewed. Comparisons of models within a single lineage are not independent, so we report 95\% confidence intervals from a cluster bootstrap over the lineages for most rows. For merges we cluster over the 11 merged models and for distillations we cluster over the base models.

\begin{table}[htbp]
\centering
\caption{$D$ and PhyloLM by model relationship.}
\label{tab:phylolm-rel}
\begin{tabular}{@{}lccc@{}}
\toprule
Relationship & $n$ pairs & \stack{Spanner: median\\{[95\% CI of the median]}} & \stack{PhyloLM: median\\{[95\% CI of the median]}} \\
\midrule
8-bit quantization & 1 & 0.020 & 0.044 \\
4-bit quantization & 1 & 0.258 & 0.272 \\
3-bit quantization & 1 & 0.528 & 0.851 \\
Fine-tunes & 59 & 0.141 [0.092, 0.243] & 0.249 [0.207, 0.375] \\
Merges vs.\ parents & 32 & 0.146 [0.034, 0.191] & 0.219 [0.152, 0.337] \\
Distillation: $d(\text{student},\text{base})$ & 11 & 0.478 [0.287, 0.690] & 0.523 [0.329, 1.161] \\
Distillation: $d(\text{student},\text{teacher})$ & 4/7\textsuperscript{a} & 1.568 [1.566, 1.568] & 0.987 [0.450, 1.732] \\
Distillation: $d(\text{base},\text{teacher})$ & 4/7\textsuperscript{a} & 1.568 [1.566, 1.568] & 0.796 [0.406, 1.744] \\
Width pruning & 3 & 1.056 & 2.843 \\
Continued pretraining & 9 & 1.441 [1.140, 1.559] & 0.430 [0.381, 0.890] \\
Pruning then distillation & 2 & 1.565 & 0.441 \\
Independent from same lab & 3,807 & 1.568 [1.566, 1.568] & 0.779 [0.651, 0.915] \\
Independent from different labs & 2,625 & 1.568 [1.568, 1.569] & 1.159 [1.003, 1.300] \\
\bottomrule
\multicolumn{4}{@{}l}{\footnotesize \textsuperscript{a} $n=4$ for $D$ because in 3 of the 7 base-student-teacher triples, the teachers are closed models.}
\end{tabular}
\end{table}

\begin{figure}[htbp]
\centering
\includegraphics[width=0.85\textwidth]{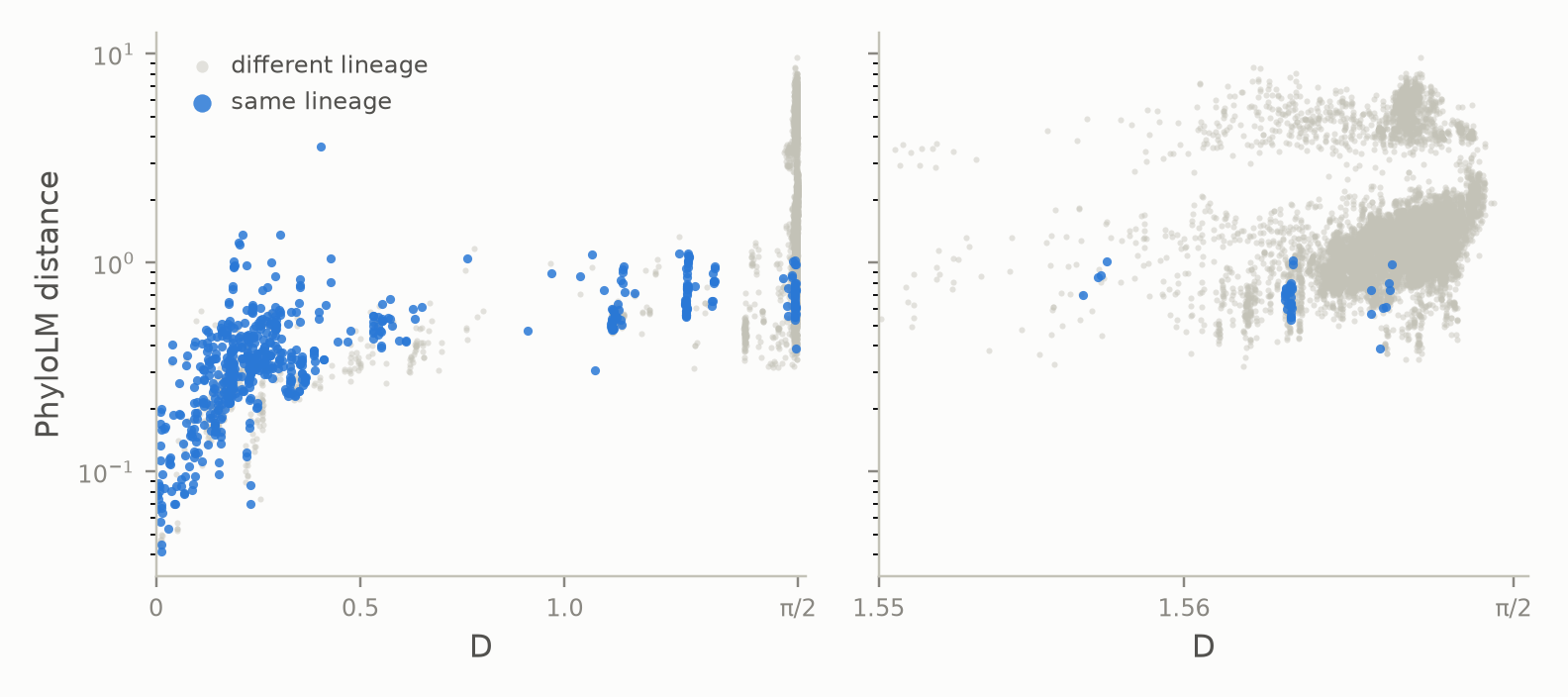}
\caption{$D$ vs.\ PhyloLM.}
\label{fig:scatter}
\end{figure}

Finally, we evaluate rank correlation between PhyloLM and $D$ across all pairs (Table~\ref{tab:phylolm-corr} and Figure~\ref{fig:scatter}). Pairs with shared models are not independent, so we report a Mantel test\citep{mantel1967detection} with 9,999 permutations for each correlation. We report 95\% confidence intervals from a bootstrap over models.

\begin{table}[htbp]
\centering
\caption{Rank correlation between $D$ and PhyloLM.}
\label{tab:phylolm-corr}
\begin{tabular}{@{}lccc@{}}
\toprule
Pairs & $n$ pairs & Spearman [95\% CI] & Mantel $p$ \\
\midrule
All pairs & 15,400 & 0.47 [0.38, 0.57] & $\le 0.0001$ \\
$D < 1.5$ & 1,233 & 0.75 [0.66, 0.81] & $\le 0.0001$ \\
$D \ge 1.5$ & 14,167 & 0.33 [0.23, 0.45] & $\le 0.0001$ \\
Within one lineage & 610 & 0.76 [0.65, 0.83] & $\le 0.0001$ \\
\bottomrule
\end{tabular}
\end{table}

We found that $D$ scores higher than PhyloLM on all five three-taxon tests, though the 95\% confidence intervals overlap. We also found that the two methods' distances are correlated (Table~\ref{tab:phylolm-corr}). As expected, the two methods disagree when weight descent is different from behavioral descent. For example, with PhyloLM, independent models from the same lab (0.779) are closer than pairs from different labs (1.159), while with $D$ the two distances are equal (1.568).

\subsection{Lineage recovery}
\label{sec:exp-lineage}

In this experiment, we evaluate $D$'s performance detecting and ordering documented relatives. We evaluate PhyloLM alongside it to show their relative strengths and to illustrate how they can be used together for lineage analysis. We ran PhyloLM with its published settings: 128 prompts from its ``math'' probe set, 32 completions of 5 tokens each per prompt at 0.7 temperature, and two seeds per model.

For this experiment, we created a set of 117 models from 34 lineages, including Qwen1.5 through Qwen3, SmolLM2, Gemma-2/3\citep{gemmateam2024gemma}, Llama-3.1/3.2, DataDecide\citep{magnusson2025datadecide}, 33 third-party Qwen fine-tunes, and 4 models we built from Qwen3-1.7B: a layer duplication, a layer deletion, and two short fine-tunes. We include 87 Qwen models in total. We only included models with descent documented in their creators' publications or model cards on the Hugging Face Hub, or that we created ourselves. See the model zoo (Appendix~\ref{app:zoo}) for details. Models with descent documented only in Hugging Face Hub metadata are evaluated in ``Spanner in the wild'' (Section~\ref{sec:exp-wild}).

We evaluated $D$ and PhyloLM's abilities to distinguish 83 known parent-child pairs, wherein the child model was trained from the parent's weights, from pairs of independent models (Table~\ref{tab:lineage-det}). We tested this in two settings: for strict detection we only consider models from different labs to be independent, and for lenient detection we add in same-lab pairs that are not documented as related to each other, such as Qwen2.5-1.5B and Qwen3-4B. We report AUROC and TPR at 1\% FPR for both cases. Comparisons of pairs within a single lineage are not independent, so we report 95\% confidence intervals from a cluster bootstrap over the 34 lineages.

\begin{table}[htbp]
\centering
\caption{Parent-child detection on the Lineage Recovery model set.}
\label{tab:lineage-det}
\begin{tabular}{@{}lccc@{}}
\toprule
Test & $n$ & Spanner [95\% CI] & PhyloLM [95\% CI] \\
\midrule
Strict detection AUROC & 2,708 pairs & \textbf{0.9994} [0.9951, 1.000] & 0.912 [0.862, 0.967] \\
Strict detection TPR at 1\% FPR & 2,708 pairs & \textbf{0.988} [0.956, 1.000] & 0.663 [0.520, 0.912] \\
Lenient detection AUROC & 6,515 pairs & \textbf{0.9955} [0.9874, 0.9998] & 0.863 [0.818, 0.925] \\
Lenient detection TPR at 1\% FPR & 6,515 pairs & \textbf{0.940} [0.882, 1.000] & 0.615 [0.488, 0.741] \\
\bottomrule
\end{tabular}
\end{table}

We also evaluate $D$ and PhyloLM with three-taxon tests\citep{nelson1991threetaxon}, like we did in the Comparison with PhyloLM experiment (Section~\ref{sec:exp-phylolm}, Table~\ref{tab:lineage-triples}). Comparisons of triples within a single lineage are not independent, so we report 95\% confidence intervals from a cluster bootstrap over the 34 lineages, as above.

\begin{table}[htbp]
\centering
\caption{Three-taxon lineage tests on the Lineage Recovery model set.}
\label{tab:lineage-triples}
\begin{tabular}{@{}lccc@{}}
\toprule
Test & $n$ & Spanner [95\% CI] & PhyloLM [95\% CI] \\
\midrule
Betweenness: ancestor and outgroup & 162 triples & 84.0\% [66.0, 97.3] & 73.5\% [56.9, 85.6] \\
Betweenness: parent of two & 165 triples & 74.5\% [57.9, 91.2] & 67.9\% [53.2, 81.5] \\
Betweenness: chain & 52 triples & 80.8\% [64.3, 92.1] & 82.7\% [70.2, 94.1] \\
Siblings closer than cousins within one lineage & 85 triples & 95.3\% [74.2, 100] & 34.1\% [17.6, 100] \\
Siblings closer than cousins across lineages & 5,053 triples & \textbf{96.4\%} [93.8, 100] & 70.6\% [57.4, 91.3] \\
\bottomrule
\end{tabular}
\end{table}

We use PhyloLM's distance matrix and $D^2$ to construct lineage trees for this set of models (Table~\ref{tab:lineage-trees}). We apply two lineage algorithms, NeighborNet\citep{bryant2004neighbornet} and NJ\citep{saitou1987neighborjoining}, to compare their results. To evaluate fit quality, we report how many ground truth families, defined as a model with all of its documented descendants, are recovered correctly, meaning that exact family is separated from other models by a single branch. We also report Holland's four-point $\delta$\citep{holland2002delta} to evaluate how well our measurements fit a tree structure. Holland's four-point $\delta$ approaches 0 as data is more tree-like and 1 as data is less tree-like. We calculate 95\% confidence intervals within lineages with cluster bootstraps over the lineages and across lineages with bootstraps over the four-point quartets.

\begin{table}[htbp]
\centering
\caption{Lineage tree recovery and four-point $\delta$.}
\label{tab:lineage-trees}
\begin{tabular}{@{}lccc@{}}
\toprule
Test & $n$ & Spanner [95\% CI] & PhyloLM [95\% CI] \\
\midrule
Families recovered by NJ & 49 families & \textbf{35 of 49} [30, 42] & 20 of 49 [13, 28] \\
Families recovered by NeighborNet & 49 families & \textbf{40 of 49} [35, 46] & 19 of 49 [13, 26] \\
Mean four-point $\delta$ within one lineage & 1,256 quartets & \textbf{0.032} [0.023, 0.055] & 0.266 [0.202, 0.321] \\
Mean four-point $\delta$ across all lineages & \stack{2,000,000\\random quartets} & \textbf{0.266} [0.265, 0.268] & 0.357 [0.355, 0.358] \\
\bottomrule
\end{tabular}
\end{table}

These results indicate that $D$ can detect lineage and performs well on our three-taxon lineage tests. In strict detection, $D$ misses only one of the 83 parent-child pairs, Llama-3.2-1B and Llama-3.1-8B, wherein Llama-3.2-1B was pruned from Llama-3.1-8B and then distilled. $D$ fits a tree structure well within a documented lineage (mean four-point $\delta$ 0.032), but it is less tree-like across lineages (mean four-point $\delta$ 0.266). PhyloLM is less tree-like than $D$ both within lineages (0.266) and across them (0.357). NeighborNet recovered about the same number of families as NJ.

\subsection{Ablations}
\label{sec:exp-ablations}

In this section, we ablate our procedure for calculating $D$ and measure performance changes in lineage detection and symmetry handling. For symmetry testing, we use the base and edited versions of Qwen3-1.7B and Llama-3.2-1B that we used in the Invariants experiment (Section~\ref{sec:exp-invariants}). For the lineage detection experiments, we used the same set of models as the Lineage Recovery experiment above (Section~\ref{sec:exp-lineage}). See the Model Zoo (Appendix~\ref{app:zoo}) for more details.

We calculate lenient detection AUROC, siblings closer than cousins and Holland's four-point $\delta$ as in the Lineage Recovery experiment. We report a summary table here (Table~\ref{tab:ablations}), but for more detail see Table~\ref{tab:ablations-full} in Appendix~\ref{app:ablations}.

\begin{table}[htbp]
\centering
\caption{Summary of ablations.}
\label{tab:ablations}
\resizebox{\textwidth}{!}{%
\begin{tabular}{@{}lccccc@{}}
\toprule
Ablation & \stack{Step\\ablated} & \stack{Lenient detection\\AUROC} & \stack{Siblings closer than cousins\\across lineages} & \stack{Mean four-point $\delta$\\within one lineage} & \stack{Largest observed $D$ between exact\\symmetries and layer duplications} \\
\midrule
Baseline & -- & 0.9955 & 0.964 & 0.0324 & 5e-6 \\
No norm folding & 1 & +0.0001 & -0.013 & -0.0014 & 0.90 \\
Raw vectors in $G$ & 3 & +0.0005 & +0.009 & -0.0003 & 0.53 \\
Unnormalized $T$ & 3 & +0.0033 & +0.025 & -0.0007 & 0.40 \\
Gram matrix instead of span & 5 & -0.0057 & -0.006 & +0.0041 & 0.29 \\
Uncentered tokens & 2 & -0.0199 & -0.105 & +0.0016 & 4e-6 \\
Uncentered traces & 3 & -0.0041 & -0.056 & +0.0002 & 4e-6 \\
No per-head $W_{QK}$ normalization & 3 & -0.0004 & -0.001 & +0.0001 & 4e-6 \\
Embeddings only (no layers) & 3 & -0.0161 & -0.079 & +0.0150 & 1e-6 \\
$k = 4{,}096$ & 4 & -0.0005 & -0.003 & +0.0000 & 5e-6 \\
$|S| = 64$, random draw 1 & 2 & -0.0013 & -0.014 & -0.0002 & -- \\
$|S| = 64$, random draw 2 & 2 & -0.0025 & -0.027 & -0.0001 & -- \\
$S$ shared with AWM's and PhyloLM's tokenizers (11,184) & 2 & +0.0010 & +0.006 & -0.0006 & -- \\
\bottomrule
\end{tabular}%
}
\end{table}

Unnormalized $T$ and some other symmetry ablations happened to improve AUROC and siblings-closer-than-cousins on this eval set, but they break scale invariance and none of the improvements were significant. Interestingly, the $k = 4{,}096$ sketches and $|S| = 64$ cases did not significantly degrade our results. This mitigates one of $D$'s biggest limitations, that $|S|$ decreases as we add tokenizers to a comparison set.

\subsection{Application: Spanner in the Wild}
\label{sec:exp-wild}

\begin{figure}[p]
\centering
\includegraphics[width=\textwidth]{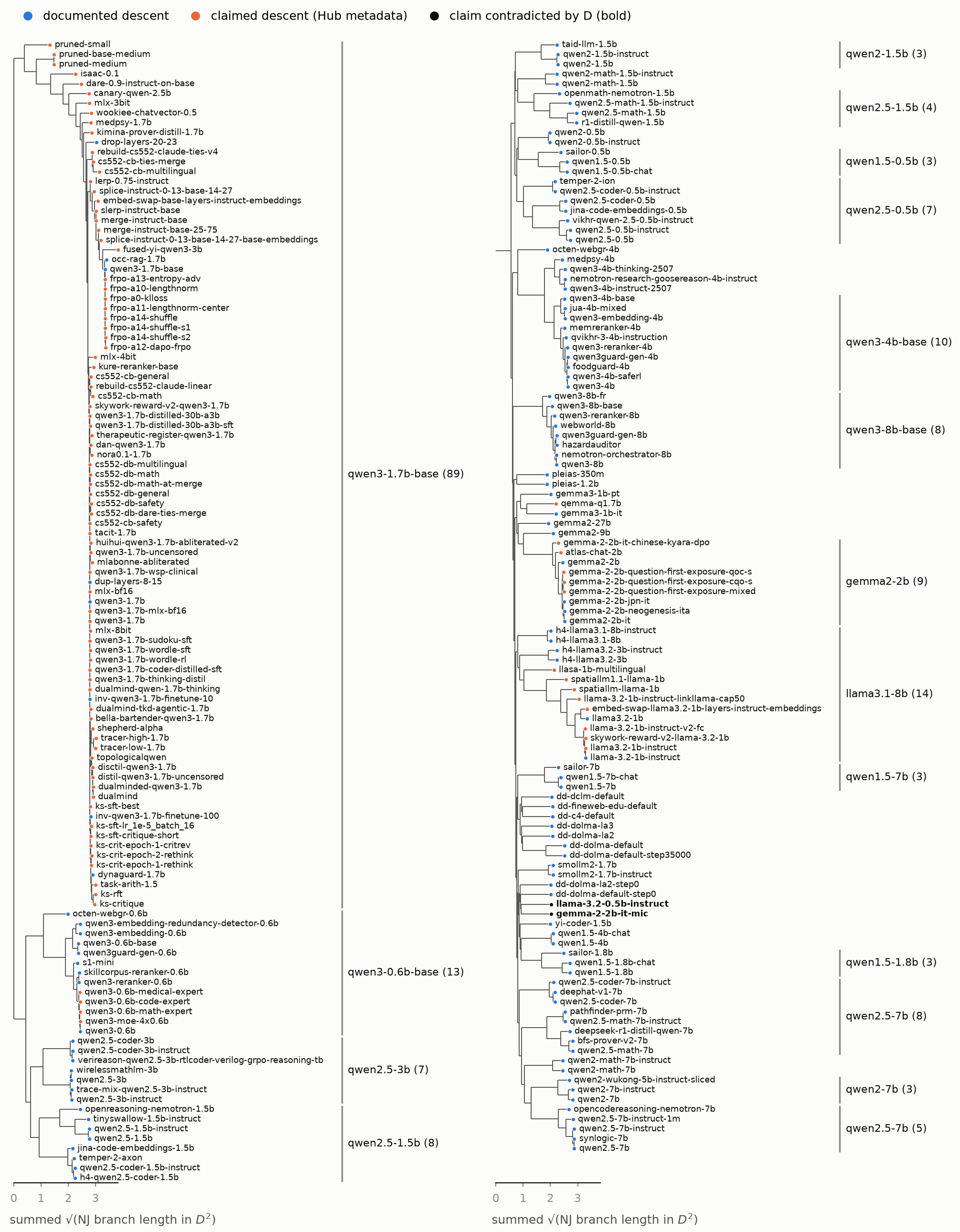}
\caption{Midpoint-rooted $D^2$ NJ Tree on 231 models.}
\label{fig:nj}
\end{figure}

In this section, we apply $D$ to an extended set of lineages combined from both official releases and community remixes. See the model zoo (Appendix~\ref{app:zoo}) for a detailed model list. See Figure~\ref{fig:nj} for a $D^2$ NJ tree of all 231 models.

Community models list parent models, but they are not verified. We audited 101 parent model labels on the Hugging Face Hub and found that $D$ disagreed with 2 of them, \nolinkurl{huggingface-mic/gemma-2-2b-it-mic} and \nolinkurl{inference-optimization/Llama-3.2-0.5B-Instruct}, which had $D = 1.5706$ to their listed parents, \nolinkurl{google/gemma-2-2b-it} and \nolinkurl{meta-llama/Llama-3.2-1B-Instruct} respectively.

We also found a handful of models where $D$ was small to the listed parent but even smaller to a closer match (Table~\ref{tab:wild}). In these cases, we followed up with median relative weight distance $\Delta_w(A,B)$ over the attention and MLP matrices $W$, $\frac{\lVert W_{\text{A}} - W_{\text{B}} \rVert_F}{\lVert W_{\text{B}} \rVert_F}$. For reference, $\Delta_w(\text{Qwen3-1.7B}, \text{Qwen3-1.7B-Base}) = 0.094$.

\begin{table}[htbp]
\centering
\caption{Hugging Face Hub parent-label disagreements.}
\label{tab:wild}
\resizebox{\textwidth}{!}{%
\begin{tabular}{@{}lllcccc@{}}
\toprule
Model & Parent listed on the Hub & Closest match & \stack{$D$(model,\\listed parent)} & \stack{$D$(model,\\closest match)} & \stack{$\Delta_w$(model,\\listed parent)} & \stack{$\Delta_w$(model,\\closest match)} \\
\midrule
FRPO (8 runs) & Qwen3-1.7B & Qwen3-1.7B-Base & 0.287 & 0.003--0.004 & 0.095 & 0.0009 \\
Qwen3-1.7B-Thinking-Distil & Disctil-Qwen3-1.7B & Qwen3-1.7B & 0.062 & 0.004 & 0.013 & 0.001 \\
Dualmind-Qwen-1.7B-Thinking & DualMinded-Qwen3-1.7B & Qwen3-1.7B-Thinking-Distil & 0.063 & 0.005 & 0.013 & 0.002 \\
Qwen3-1.7B-Base-pruned-medium & Qwen3-1.7B-Base & Qwen3-1.7B & 1.056 & 1.034 & 0.097 & 0.001 \\
\bottomrule
\end{tabular}%
}
\end{table}

\nolinkurl{rookshanks/Qwen3-1.7B-Base-pruned-medium} is an exact copy of \nolinkurl{rookshanks/Qwen3-1.7B-medium-pruned}, which is a pruned Qwen3-1.7B. Its remaining weights match Qwen3-1.7B to $\Delta_w = 0.001$.

We cannot confidently say that these listings are incorrect, but these results prevent us from verifying their provenance. We did not have to look very hard to find these examples, so a larger sweep of LLMs on the Hugging Face Hub would likely catch more. Further, such a sweep may detect matches for models that were uploaded without a parent tag.

\subsection{Performance and Cost}
\label{sec:exp-cost}

We timed $D$, AWM, PhyloLM, Trace, and Subspace Alignment on comparisons of Qwen3-0.6B, Qwen3-1.7B, Qwen3-4B, and Qwen3-8B and their base models on a Hugging Face l40sx1 (NVIDIA L40S, 8 vCPUs, \$1.80/h). We ran each pair three times on fresh instances. We calculate how much time and cost it would take to compare 1,000 models of each size with each method (Table~\ref{tab:cost}).

\begin{table}[htbp]
\centering
\caption{Performance and cost for 1,000 models.}
\label{tab:cost}
\begin{tabular}{@{}lcccc@{}}
\toprule
Method & 0.6B & 1.7B & 4B & 8B \\
\midrule
Spanner ($D$) & 2.8 h (\$5) & 4.5 h (\$8) & 6.9 h (\$12) & 16.7 h (\$30) \\
AWM & 60 days (\$2,583) & 64 days (\$2,782) & 71 days (\$3,088) & 86 days (\$3,704) \\
PhyloLM & 5.8 h (\$11) & 6.5 h (\$12) & 9.4 h (\$17) & 11.8 h (\$21) \\
Biometrics: Trace & 3.3 h (\$6) & 3.8 h (\$7) & 5.2 h (\$9) & 7.2 h (\$13) \\
Biometrics: Subspace & 62 days (\$2,690) & 65 days (\$2,792) & 92 days (\$3,986) & 99 days (\$4,296) \\
\bottomrule
\end{tabular}
\end{table}

We excluded model download times from these calculations. We timed PhyloLM with our own optimized sampler and used the author's default GPU float16 SVD for Biometrics Subspace.

$D$'s pairwise distance comparison takes milliseconds because we sketch each model once. Those fast pairwise distance comparisons allow Hub-scale lineage analysis. Trace and PhyloLM cost about the same, but $D$ outperforms those methods. See Table~\ref{tab:cost-full} in Appendix~\ref{app:cost} for an extended performance and cost table.

\section{Limitations}
\label{sec:limitations}

\begin{itemize}
  \item $D$ requires a shared token set $S$, which may shrink with additional tokenizers. Distances are not directly comparable across studies unless they share a fixed $S$.
  \item Sketching $T$ while calculating $D$ adds a small amount of sketch noise and necessitates re-sketching when the token set changes.
  \item $D$ ignores model differences in biases, norms, nonlinearities, layer order, and individual attention head structure. $D$ also ignores descent direction.
  \item $D$ saturates near $\frac{\pi}{2}$. Sketch bias and weak overlap in weights between independently trained models can cause $D$ to decrease to slightly below $\frac{\pi}{2}$, so values near saturation are not proof of shared weight lineage.
  \item Model pruning and embedding swaps can increase $D$ despite shared weights.
  \item $D$ cannot detect distillation, so it should be paired with PhyloLM or another behavioral lineage measurement.
  \item We did not test $D$ against an adversary. An attacker could potentially hide a model's true lineage.
  \item We designed $D$ for attention and MLP sublayers. We will need to create new invariants for other architectures.
  \item Most of the models we used to evaluate $D$ are small by frontier standards. It is possible that larger models would require changes to this method, such as increasing the size of our CountSketches.
\end{itemize}

\section{Discussion}
\label{sec:discussion}

We hope to soon apply Spanner to various tasks in LLM engineering and safety. Does $D$ measure compatibility for merging models? Can $D$ predict retained performance after quantization? Under fine-tuning, at what $D$ do the base model's alignment safeguards stop working\citep{qi2024finetuning}? If we teach model $X$ a new fact to make $X'$, what is $D(X, X')$?

As we have shown, weight-based and behavior-based measurements of model similarity can disagree. Pairs such as independent models from the same lab and distilled models are closer in their behavior than their weights, whereas merged and spliced models can be closer to their parents' weights than their behavior. These differences indicate that analyses of model lineage should pair weight analysis with behavioral analysis just like biologists analyze both genotype and phenotype.

$D$ is a functional training clock and $D^2$ correlates closely with cumulative learning rate. We can use these properties to monitor training runs and estimate cumulative learning rate between pairs of models. $D$ always increases to $\frac{\pi}{2}$ after enough training, but even in its saturated range it correlates with behavioral distance. $D$ has a Spearman correlation of 0.33 with PhyloLM for pairs with $D \ge 1.5$. With more work, we may find a different derivation of $D$ that achieves better dynamic range while comparing near-independent models.

$D$ enables comparing large numbers of models across tokenizers with reasonable performance and cost. $D$'s performance enables auditing model lineages within large collections such as the Hugging Face Hub. Using both PhyloLM and $D$, we could conceivably verify each model's listed parents.

LLM lineages are not trees. LLMs can merge, splice, or assemble into MoE models in a way that is more like hybridization or horizontal gene transfer. As LLM adoption and remixing continue, Spanner and other measurements like it will be crucial for tracking model evolution and adaptive radiation.

\bibliographystyle{unsrtnat-etal}
\bibliography{../related_work/refs}

\clearpage
\appendix

\section{Model zoo}
\label{app:zoo}

For a full list of models with each model's Hugging Face repository and pinned commit, see \href{https://github.com/matthew-ritch/spanner/blob/main/results/model_zoo/model_zoo.csv}{\nolinkurl{results/model_zoo/model_zoo.csv}} in our repository (Table~\ref{tab:zoo}).

\begin{table}[htbp]
\centering
\caption{Model zoo summary.}
\label{tab:zoo}
\begin{tabularx}{\textwidth}{@{} l c >{\raggedright\arraybackslash}X @{}}
\toprule
Experiment & Models & Contents \\
\midrule
Invariants & 26 & Qwen3-1.7B and Llama-3.2-1B (Base and Instruct), MLX quantizations, a Qwen3-0.6B MoE and its experts, and our builds and fine-tunes \\
Sketch noise & 9 & Qwen3-1.7B, SmolLM2-1.7B, Llama-3.2-1B and Gemma-2-2B (Base and Instruct), and Qwen2.5-1.5B-Instruct \\
Training clock & 94 & Pythia-410M\citep{biderman2023pythia} \\
Lineage recovery & 117 & See \href{https://github.com/matthew-ritch/spanner/blob/main/results/model_zoo/model_zoo.csv}{full list} \\
Comparison with AWM & 94 & AWM's benchmark set\citep{zeng2026awm} \\
Comparison with LLM Biometrics & 121 & LLM Biometrics' benchmark set\citep{chen2026who} and AWM's benchmark set \\
Comparison with PhyloLM & 316 & See \href{https://github.com/matthew-ritch/spanner/blob/main/results/model_zoo/model_zoo.csv}{full list} \\
Application: Spanner ``in the wild'' & 231 & See \href{https://github.com/matthew-ritch/spanner/blob/main/results/model_zoo/model_zoo.csv}{full list} \\
Performance and Cost & 8 & Qwen3-0.6B, 1.7B, 4B and 8B (Base and Instruct) \\
\bottomrule
\end{tabularx}
\end{table}

For more details, see the full model zoo in our repository at \url{https://github.com/matthew-ritch/spanner/blob/main/results/model_zoo/model_zoo.csv}.

\section{Token Sets}
\label{app:tokens}

$S$, the token set $D$ uses, is the intersection of the compared models' token sets and thus varies by experiment (Table~\ref{tab:tokens}).

\begin{table}[htbp]
\centering
\caption{Token set sizes by experiment.}
\label{tab:tokens}
\begin{tabularx}{\textwidth}{@{} >{\raggedright\arraybackslash}X r @{}}
\toprule
Experiment & $|S|$ \\
\midrule
Invariants, Sketch noise, Lineage recovery, Ablations, Spanner ``in the wild'', Performance and Cost & 24,554 \\
Training clock & 50,252 \\
Comparison with AWM & 13,081 \\
Comparison with LLM Biometrics & 16,492 \\
Comparison with PhyloLM & 13,586 \\
\bottomrule
\end{tabularx}
\end{table}

\section{Sketching Fingerprint Matrices}
\label{app:sketch}

\begin{itemize}
  \item Define $a_c$ as column $c$ of $E F$ or $U F$
  \item Define $b_c$ as column $c$ of $E$ or $U$
  \item Then each $T = \sum_c a_c b_c^\top$ over the $d$ columns $c$.
\end{itemize}

For each invariant type and token set, let $\text{sketch}_1, \text{sketch}_2 : \mathbb{R}^{|S|} \to \mathbb{R}^k$ be two independent CountSketches with $k = 16{,}384$ with hashes $h_m : S \to \{0, \dots, k-1\}$ and signs $s_m : S \to \{-1,+1\}$, where $S$ is taken in a fixed order such as sorted by detokenized strings. Thus $\text{sketch}_m(x)_r = \sum_{i:\,h_m(i) = r} s_m(i)\,x_i$. We create them once and use them for each model, so all fingerprints are converted to the same $\mathbb{R}^k$.

The TensorSketch of a rank-one matrix $a b^\top$ is the circular convolution of the two sketches $\text{sketch}_1(a) \ast \text{sketch}_2(b)$. This is a TensorSketch of $(i,j)$ with hash $\bigl(h_1(i) + h_2(j)\bigr) \bmod k$ and sign $s_1(i)\, s_2(j)$. The sketches are linear, so we can apply the convolution theorem\citep{pham2013fast} and sum the rank-one terms of the invariants in the Fourier domain and then invert to calculate:
\[
\text{sketch}(T) = \text{IFFT}\Bigl(\sum_{c=1}^{d} \text{FFT}\bigl(\text{sketch}_1(a_c)\bigr) \odot \text{FFT}\bigl(\text{sketch}_2(b_c)\bigr)\Bigr)
\]

$\mathbb{E}\langle \text{sketch}(T), \text{sketch}(T')\rangle = \langle T, T' \rangle_F$ with variance $O(\frac{1}{k})$, so we can compare sublayers without forming a $|S| \times |S|$ matrix. Sketching and transforming $E$ and $U$ depends only on the model, so it is done once per model, at a cost of $O(d|S| + d k \log k)$. Then, it costs $O(k d^2)$ to produce $T$ by matrix multiplying those with $F$ in the Fourier domain.

We need to ensure that the error it adds is small relative to our measurements. We test this by comparing exact and sketched $D$ on 7 pairs of models (Table~\ref{tab:sketch-exact}).

\begin{table}[htbp]
\centering
\caption{Exact vs.\ sketched $D$.}
\label{tab:sketch-exact}
\begin{tabular}{@{}lcccc@{}}
\toprule
Pair & Exact $D$ & Sketched $D$ & Error & Predicted bias ($\cos\theta$) \\
\midrule
Qwen3-1.7B Base / Instruct & 0.28706 & 0.28691 & $-1.6 \times 10^{-4}$ & -- \\
SmolLM2-1.7B / -Instruct & 0.24299 & 0.24302 & $+3.1 \times 10^{-5}$ & -- \\
Llama-3.2-1B / -Instruct & 0.48569 & 0.48575 & $+5.5 \times 10^{-5}$ & -- \\
Gemma-2-2B / -it & 0.19970 & 0.19972 & $+1.5 \times 10^{-5}$ & -- \\
Qwen3-1.7B-Base / SmolLM2-1.7B & 1.56945 & 1.56879 & $-6.6 \times 10^{-4}$ & $6.6 \times 10^{-4}$ \\
Qwen2.5-1.5B-Instruct / Qwen3-1.7B & 1.56117 & 1.56045 & $-7.2 \times 10^{-4}$ & $7.1 \times 10^{-4}$ \\
Gemma-2-2B-it / Llama-3.2-1B-Instruct & 1.56538 & 1.56485 & $-5.3 \times 10^{-4}$ & $5.2 \times 10^{-4}$ \\
\bottomrule
\end{tabular}
\end{table}

Sketching the models into $k$ dimensions makes unrelated models overlap slightly by chance, so $D$ will be a little below $\frac{\pi}{2}$. The magnitude of this bias can be computed from the models' layer counts and $k$.
\begin{itemize}
  \item We concatenate each sublayer's invariant sketches to make their fingerprint vectors. Attention sublayers have 2 invariant sketches ($O$, $Q$), so $f_{\text{attn}} \in \mathbb{R}^{2k}$, and MLP sublayers have 3 invariant sketches ($M$, $G_{\text{in}}$, $G_{\text{out}}$), so $f_{\text{mlp}} \in \mathbb{R}^{3k}$.
  \item For two IID subspaces $A$ and $B$ of dimensions $r_A$ and $r_B$ in an $n$-dimensional space, the expected Frobenius inner product of their orthogonal projection matrices $P_A$ and $P_B$ is $\mathbb{E}\langle P_A, P_B \rangle = \frac{r_A r_B}{n}$.
  \item We assume that two unrelated models' sketched spans behave like random subspaces. Use $r$ to denote the rank of $K_t$, usually the number of sublayers of type $t$. Then, using the expression for $\cos\theta$ in step 5 of $D$'s procedure, we can calculate the predicted bias:
  \[
    b = \frac{\frac{1}{2k} r_{A,\text{attn}} r_{B,\text{attn}} + \frac{1}{3k} r_{A,\text{mlp}} r_{B,\text{mlp}}}{\sqrt{(r_{A,\text{attn}} + r_{A,\text{mlp}})(r_{B,\text{attn}} + r_{B,\text{mlp}})}}.
  \]
  For example, two 28-layer models with $k = 16{,}384$ give $b = 7.1 \times 10^{-4}$.
  \item Thus, we expect that:
  \begin{enumerate}
    \item for an unrelated pair, $\cos\theta_{\text{sketch}} = \cos\theta_{\text{exact}} + b$;
    \item doubling $k$ should halve $b$, so $\cos\theta_{16k} - \cos\theta_{32k} = b_{16k}/2$.
  \end{enumerate}
\end{itemize}

We now test these expectations empirically (Table~\ref{tab:sketch-bias}).

\begin{table}[htbp]
\centering
\caption{Empirical vs.\ predicted sketch bias.}
\label{tab:sketch-bias}
\resizebox{\textwidth}{!}{%
\begin{tabular}{@{}lcccc@{}}
\toprule
Test & Pairs & Measurement & Observed & Predicted \\
\midrule
Random initializations, 16k & 1 & $\cos\theta_{16k}$ (exact is $\sim$0) & $3.87 \times 10^{-4}$ & $4.07 \times 10^{-4}$ \\
Random initializations, 32k & 1 & $\cos\theta_{32k}$ & $1.95 \times 10^{-4}$ & $2.03 \times 10^{-4}$ \\
Qwen3-1.7B-Base / SmolLM2-1.7B & 1 & $\cos\theta_{16k} - \cos\theta_{\text{exact}}$ & $6.60 \times 10^{-4}$ & $6.59 \times 10^{-4}$ \\
Qwen2.5-1.5B-Instruct / Qwen3-1.7B & 1 & same & $7.19 \times 10^{-4}$ & $7.12 \times 10^{-4}$ \\
Gemma-2-2B-it / Llama-3.2-1B-Instruct & 1 & same & $5.34 \times 10^{-4}$ & $5.19 \times 10^{-4}$ \\
AWM's independent pairs & 90 & mean $\cos\theta_{16k} - \cos\theta_{32k}$ & $4.62 \times 10^{-4}$ & $4.58 \times 10^{-4}$ (corr.\ across pairs 0.82) \\
Collection, different labs & 80 & $(\cos\theta_{16k} - \cos\theta_{32k}) \,/\, \tfrac{b_{16k}}{2}$ & median 1.10 (IQR 1.01--1.16) & 1 \\
\bottomrule
\end{tabular}%
}
\end{table}

\section{Performance and Cost Comparison}
\label{app:cost}

\begin{table}[htbp]
\centering
\caption{Extended performance and cost comparison.}
\label{tab:cost-full}
\resizebox{\textwidth}{!}{%
\begin{tabular}{@{}lcllll@{}}
\toprule
Method & Size & Per model & Per pair & Cache per model & Total, $N = 1{,}000$ \\
\midrule
Spanner ($D$) & 0.6B & 8.7 s (8.2 s--9.2 s) & 2.8 ms (2.6 ms--3.6 ms) & 37 MB & 2.8 h (\$5) \\
 & 1.7B & 15 s (14 s--16 s) & 2.7 ms (2.6 ms--3.4 ms) & 37 MB & 4.5 h (\$8) \\
 & 4B & 23 s (20 s--24 s) & 4.3 ms (4.0 ms--4.3 ms) & 48 MB & 6.9 h (\$12) \\
 & 8B & 58 s (56 s--64 s) & 3.9 ms (3.9 ms--6.2 ms) & 48 MB & 16.7 h (\$30) \\
\addlinespace
AWM & 0.6B & none & 10 s (10 s--11 s) & none (re-reads 1.3 GB of weights) & 60 days (\$2,583) \\
 & 1.7B & none & 11 s & none (re-reads 3.8 GB of weights) & 64 days (\$2,782) \\
 & 4B & none & 12 s (12 s--13 s) & none (re-reads 8.0 GB of weights) & 71 days (\$3,088) \\
 & 8B & none & 15 s (14 s--21 s) & none (re-reads 16 GB of weights) & 86 days (\$3,704) \\
\addlinespace
PhyloLM & 0.6B & 21 s (17 s--24 s) & 690 $\mu$s (678 $\mu$s--727 $\mu$s) & 319 kB & 5.8 h (\$11) \\
 & 1.7B & 23 s (19 s--27 s) & 521 $\mu$s (520 $\mu$s--526 $\mu$s) & 319 kB & 6.5 h (\$12) \\
 & 4B & 34 s (30 s--38 s) & 562 $\mu$s (512 $\mu$s--761 $\mu$s) & 319 kB & 9.4 h (\$17) \\
 & 8B & 42 s (39 s--46 s) & 521 $\mu$s (468 $\mu$s--636 $\mu$s) & 322 kB & 11.8 h (\$21) \\
\addlinespace
Biometrics: Trace & 0.6B & 9.8 s (5.8 s--14 s) & 4.3 ms (4.2 ms--4.3 ms) & 2.5 kB & 3.3 h (\$6) \\
 & 1.7B & 12 s (7.6 s--16 s) & 4.3 ms (4.2 ms--4.3 ms) & 2.5 kB & 3.8 h (\$7) \\
 & 4B & 17 s (12 s--21 s) & 4.3 ms (4.2 ms--4.3 ms) & 2.7 kB & 5.2 h (\$9) \\
 & 8B & 24 s (20 s--28 s) & 4.2 ms (4.2 ms--4.3 ms) & 2.7 kB & 7.2 h (\$13) \\
\addlinespace
Biometrics: subspace & 0.6B & 17 s & 11 s & 705 MB & 62 days (\$2,690) \\
 & 1.7B & 36 s & 11 s & 1.2 GB & 65 days (\$2,792) \\
 & 4B & 80 s (79 s--82 s) & 16 s (14 s--16 s) & 2.3 GB & 92 days (\$3,986) \\
 & 8B & 7.7 min & 16 s (16 s--17 s) & 2.9 GB & 99 days (\$4,296) \\
\bottomrule
\end{tabular}%
}
\end{table}

\section{Ablations}
\label{app:ablations}

\begin{longtable}{@{}>{\raggedright\arraybackslash}p{4.0cm}cccccc@{}}
\caption{Full ablation results.}
\label{tab:ablations-full} \\
\toprule
Ablation & \stack{Step} & \stack{Lenient\\AUROC} & \stack{Strict\\AUROC} & \stack{Siblings\\(across)} & \stack{Mean $\delta$\\(within)} & \stack{Max $D$,\\sym.} \\
\midrule
\endfirsthead
\toprule
Ablation & \stack{Step} & \stack{Lenient\\AUROC} & \stack{Strict\\AUROC} & \stack{Siblings\\(across)} & \stack{Mean $\delta$\\(within)} & \stack{Max $D$,\\sym.} \\
\midrule
\endhead
\bottomrule
\multicolumn{7}{@{}l}{\footnotesize $^*$indicates a 95\% CI that excludes 0.}
\endfoot
Baseline & -- & \stack{0.9955\\{[0.9874, 0.9998]}} & \stack{0.9994\\{[0.9951, 1.0000]}} & \stack{0.964\\{[0.938, 1.000]}} & \stack{0.0324\\{[0.0228, 0.0549]}} & 5e-6 \\
\addlinespace
\multicolumn{7}{@{}l}{\textbf{Symmetry handling}} \\
No norm folding & 1 & \stack{+0.0001\\{[-0.0009, +0.0015]}} & \stack{+0.0001\\{[-0.0005, +0.0008]}} & \stack{-0.013\\{[-0.033, +0.004]}} & \stack{-0.0014\\{[-0.0034, +0.0109]}} & 0.90 \\
No unit vectors in G & 3 & \stack{+0.0005\\{[-0.0002, +0.0023]}} & \stack{+0.0000\\{[-0.0004, +0.0015]}} & \stack{+0.009\\{[-0.001, +0.021]}} & \stack{-0.0003\\{[-0.0018, +0.0000]}} & 0.53 \\
No per-T normalization & 3 & \stack{+0.0033\\{[-0.0000, +0.0110]}} & \stack{+0.0006\\{[+0.0000, +0.0049]}} & \stack{+0.025\\{[+0.000, +0.048]}} & \stack{-0.0007\\{[-0.0031, +0.0041]}} & 0.40 \\
Gram matrix in place of the projector & 5 & \stack{-0.0057\\{[-0.0190, +0.0009]}} & \stack{-0.0056\\{[-0.0179, +0.0000]}} & \stack{-0.006\\{[-0.049, +0.013]}} & \stack{+0.0041\\{[-0.0082, +0.0220]}} & 0.29 \\
\addlinespace
\multicolumn{7}{@{}l}{\textbf{Other steps}} \\
No token centering & 2 & \stack{-0.0199\\{[-0.0446, +0.0000]}} & \stack{-0.0222\\{[-0.0504, +0.0000]}} & \stack{-0.105\\{[-0.169, +0.000]}} & \stack{+0.0016\\{[-0.0021, +0.0142]}} & 4e-6 \\
No trace-centering & 3 & \stack{-0.0041\\{[-0.0114, +0.0000]}} & \stack{-0.0025\\{[-0.0128, +0.0000]}} & \stack{-0.056\\{[-0.109, +0.000]}} & \stack{+0.0002\\{[-0.0012, +0.0014]}} & 4e-6 \\
No per-head $W_{QK}$ normalization & 3 & \stack{-0.0004\\{[-0.0015, +0.0001]}} & \stack{-0.0002\\{[-0.0013, +0.0000]}} & \stack{-0.001\\{[-0.007, +0.004]}} & \stack{+0.0001\\{[-0.0000, +0.0004]}} & 4e-6 \\
\addlinespace
\multicolumn{7}{@{}l}{\textbf{Invariants}} \\
Drop O & 3 & \stack{+0.0013\\{[-0.0024, +0.0078]}} & \stack{+0.0003\\{[-0.0001, +0.0042]}} & \stack{-0.023\\{[-0.054, +0.037]}} & \stack{+0.0019\\{[-0.0041, +0.0053]}} & 5e-6 \\
Drop Q & 3 & \stack{-0.0033\\{[-0.0144, +0.0006]}} & \stack{-0.0038\\{[-0.0150, +0.0000]}} & \stack{-0.009\\{[-0.052, +0.015]}} & \stack{+0.0001\\{[-0.0018, +0.0039]}} & 5e-6 \\
Drop M & 3 & \stack{+0.0005\\{[-0.0006, +0.0028]}} & \stack{-0.0001\\{[-0.0007, +0.0001]}} & \stack{-0.009\\{[-0.021, +0.001]}} & \stack{+0.0004\\{[-0.0003, +0.0036]}} & 5e-6 \\
Drop $G_{\text{in}}$ & 3 & \stack{+0.0007\\{[-0.0009, +0.0048]}} & \stack{-0.0000\\{[-0.0007, +0.0031]}} & \stack{+0.004\\{[-0.021, +0.030]}} & \stack{-0.0008\\{[-0.0033, +0.0006]}} & 5e-6 \\
Drop $G_{\text{out}}$ & 3 & \stack{-0.0028\\{[-0.0062, +0.0002]}} & \stack{-0.0008\\{[-0.0042, +0.0001]}} & \stack{-0.004\\{[-0.039, +0.030]}} & \stack{+0.0015\\{[-0.0012, +0.0053]}} & 4e-6 \\
Attention only & 5 & \stack{+0.0021\\{[-0.0002, +0.0072]}} & \stack{+0.0004\\{[+0.0000, +0.0034]}} & \stack{+0.011\\{[-0.019, +0.036]}} & \stack{+0.0050\\{[-0.0008, +0.0207]}} & 3e-6 \\
MLP only & 5 & \stack{-0.0027\\{[-0.0085, +0.0000]}} & \stack{-0.0020\\{[-0.0079, +0.0000]}} & \stack{-0.016\\{[-0.035, +0.007]}} & \stack{-0.0025\\{[-0.0052, +0.0013]}} & 6e-6 \\
Embeddings only (no layers) & 3 & \stack{-0.0161$^*$\\{[-0.0351, -0.0026]}} & \stack{-0.0123\\{[-0.0348, +0.0000]}} & \stack{-0.079\\{[-0.167, +0.000]}} & \stack{+0.0150\\{[-0.0168, +0.0465]}} & 1e-6 \\
\addlinespace
\multicolumn{7}{@{}l}{\textbf{Sketch}} \\
$k = 4{,}096$ & 4 & \stack{-0.0005\\{[-0.0021, +0.0001]}} & \stack{-0.0004\\{[-0.0017, +0.0000]}} & \stack{-0.003\\{[-0.020, +0.006]}} & \stack{+0.0000\\{[-0.0001, +0.0001]}} & 5e-6 \\
$k = 8{,}192$ & 4 & \stack{-0.0003\\{[-0.0013, +0.0001]}} & \stack{-0.0002\\{[-0.0010, +0.0000]}} & \stack{+0.001\\{[-0.003, +0.006]}} & \stack{+0.0000\\{[+0.0000, +0.0002]}} & 5e-6 \\
$k = 32{,}768$ & 4 & \stack{+0.0001\\{[-0.0000, +0.0003]}} & \stack{+0.0000\\{[-0.0000, +0.0001]}} & \stack{-0.000\\{[-0.006, +0.005]}} & \stack{-0.0000\\{[-0.0001, +0.0003]}} & 5e-6 \\
Hash seed 5000 & 4 & \stack{+0.0001\\{[-0.0001, +0.0005]}} & \stack{+0.0000\\{[-0.0000, +0.0005]}} & \stack{+0.001\\{[-0.001, +0.004]}} & \stack{-0.0000\\{[-0.0001, +0.0003]}} & -- \\
Hash seed 6000 & 4 & \stack{-0.0001\\{[-0.0003, +0.0000]}} & \stack{-0.0000\\{[-0.0002, +0.0001]}} & \stack{+0.001\\{[-0.002, +0.003]}} & \stack{-0.0001\\{[-0.0003, +0.0002]}} & -- \\
Eigenvalue cutoff 1e-14 & 5 & \stack{+0.0000\\{[+0.0000, +0.0000]}} & \stack{+0.0000\\{[+0.0000, +0.0000]}} & \stack{+0.000\\{[+0.000, +0.000]}} & \stack{+0.0000\\{[+0.0000, +0.0000]}} & 5e-6 \\
Eigenvalue cutoff 1e-06 & 5 & \stack{+0.0000\\{[+0.0000, +0.0000]}} & \stack{+0.0000\\{[+0.0000, +0.0000]}} & \stack{+0.000\\{[+0.000, +0.000]}} & \stack{+0.0000\\{[+0.0000, +0.0000]}} & 5e-6 \\
Eigenvalue cutoff 0.001 & 5 & \stack{+0.0000\\{[+0.0000, +0.0000]}} & \stack{+0.0000\\{[+0.0000, +0.0000]}} & \stack{+0.000\\{[+0.000, +0.000]}} & \stack{+0.0000\\{[+0.0000, +0.0000]}} & 5e-6 \\
\addlinespace
\multicolumn{7}{@{}l}{\textbf{Token set $S$}} \\
$|S| = 64$, random draw 1 & 2 & \stack{-0.0013\\{[-0.0069, +0.0003]}} & \stack{-0.0025\\{[-0.0095, +0.0000]}} & \stack{-0.014\\{[-0.050, +0.025]}} & \stack{-0.0002\\{[-0.0052, +0.0016]}} & -- \\
$|S| = 64$, random draw 2 & 2 & \stack{-0.0025\\{[-0.0098, +0.0001]}} & \stack{-0.0032\\{[-0.0122, +0.0000]}} & \stack{-0.027\\{[-0.065, +0.014]}} & \stack{-0.0001\\{[-0.0032, +0.0025]}} & -- \\
$|S| = 256$, random draw 1 & 2 & \stack{+0.0004\\{[-0.0007, +0.0017]}} & \stack{-0.0007\\{[-0.0026, +0.0000]}} & \stack{-0.014\\{[-0.037, +0.000]}} & \stack{+0.0001\\{[-0.0002, +0.0006]}} & -- \\
$|S| = 256$, random draw 2 & 2 & \stack{-0.0011\\{[-0.0057, +0.0002]}} & \stack{-0.0021\\{[-0.0081, +0.0000]}} & \stack{-0.013\\{[-0.047, +0.008]}} & \stack{-0.0004$^*$\\{[-0.0023, -0.0001]}} & -- \\
$|S| = 1{,}024$, random draw 1 & 2 & \stack{+0.0004\\{[-0.0001, +0.0018]}} & \stack{-0.0002\\{[-0.0008, +0.0003]}} & \stack{-0.005\\{[-0.022, +0.003]}} & \stack{-0.0002\\{[-0.0010, -0.0000]}} & -- \\
$|S| = 1{,}024$, random draw 2 & 2 & \stack{+0.0002\\{[-0.0001, +0.0006]}} & \stack{-0.0001\\{[-0.0003, +0.0001]}} & \stack{+0.002\\{[-0.001, +0.005]}} & \stack{-0.0003$^*$\\{[-0.0009, -0.0001]}} & -- \\
$|S| = 4{,}096$, random draw 1 & 2 & \stack{-0.0001\\{[-0.0004, +0.0000]}} & \stack{-0.0000\\{[-0.0002, +0.0001]}} & \stack{-0.001\\{[-0.007, +0.004]}} & \stack{-0.0000\\{[-0.0001, +0.0007]}} & -- \\
$|S| = 4{,}096$, random draw 2 & 2 & \stack{+0.0000\\{[-0.0002, +0.0002]}} & \stack{-0.0001\\{[-0.0005, +0.0000]}} & \stack{+0.003\\{[-0.002, +0.009]}} & \stack{-0.0002\\{[-0.0004, +0.0001]}} & -- \\
$|S| = 1{,}024$, most frequent & 2 & \stack{-0.0060\\{[-0.0170, +0.0004]}} & \stack{-0.0022\\{[-0.0105, +0.0000]}} & \stack{+0.006\\{[-0.034, +0.032]}} & \stack{+0.0051$^*$\\{[+0.0014, +0.0143]}} & -- \\
$|S| = 4{,}096$, most frequent & 2 & \stack{-0.0016\\{[-0.0060, +0.0005]}} & \stack{-0.0004\\{[-0.0022, +0.0003]}} & \stack{+0.017\\{[-0.006, +0.037]}} & \stack{+0.0025$^*$\\{[+0.0005, +0.0093]}} & -- \\
S shared with AWM's and PhyloLM's tokenizers (11,184) & 2 & \stack{+0.0010\\{[+0.0000, +0.0033]}} & \stack{+0.0002\\{[+0.0000, +0.0015]}} & \stack{+0.006\\{[-0.006, +0.019]}} & \stack{-0.0006\\{[-0.0028, -0.0000]}} & -- \\
\end{longtable}

\end{document}